\documentclass[runningheads]{llncs}

\usepackage{eccv}

\usepackage{eccvabbrv}
\usepackage{caption}

\usepackage{times}
\usepackage{epsfig}
\usepackage{graphicx}
\usepackage{amsmath}
\usepackage{amssymb}
\usepackage{bbm}
\usepackage{booktabs}
\usepackage{multirow}
\usepackage{mathrsfs}
\usepackage{nicefrac}       
\usepackage[accsupp]{axessibility}  

\usepackage{hyperref}
\usepackage{hypcap}
\usepackage{orcidlink}

\usepackage[dvipsnames]{xcolor}
\usepackage{comment}
\usepackage{capt-of}
\definecolor{green}{rgb}{0, 0.5, 0}
\definecolor{orange}{rgb}{0.8, 0.6, 0.2}
\definecolor{red}{rgb}{1.0, 0.0, 0.0}
\definecolor{teal}{rgb}{0.0, 0.4, 0.4}
\definecolor{purple}{rgb}{0.65,0,0.65}
\definecolor{saffron}{rgb}{0.95,0.75,0.2}
\definecolor{turquoise}{rgb}{0.0,0.5,0.5}
\definecolor{black}{rgb}{0.0, 0.0, 0.0}
\definecolor{gray}{rgb}{0.5, 0.5, 0.5}

\newcommand{\kai}[1]{{\color{black}#1}}
\newcommand{\ym}[1]{{\color{black}#1}}
\newcommand{\rz}[1]{{\color{black}#1}}
\newcommand{\rzz}[1]{{\color{black}#1}}
\newcommand{\pg}[1]{{\color{black}#1}}

\newcommand{\nv}[1]{\mathbf{#1}}
\newcommand{\mysubsubsection}[1]{\vspace{3pt}\noindent {\bf #1}:}
\newcommand{\first}[1]{\textcolor{cyan}{#1}}
\newcommand{\second}[1]{\textcolor{orange}{#1}}

\newcommand{\fg}[1]{{\color{black}#1}}

\begin{document}

\title{DPA-Net: Structured 3D Abstraction from Sparse Views via Differentiable Primitive Assembly} 

\titlerunning{DPA-Net}

\author{Fenggen Yu\inst{1,2}\thanks{Work carried out during internship at Amazon.}\orcidlink{0000-0003-1591-4668} \and
Yiming Qian\inst{1} \and Xu Zhang\inst{1} \and
Francisca Gil-Ureta\inst{1} \and \\ Brian Jackson\inst{1} \and Eric Bennett\inst{1} \and Hao Zhang\inst{1,2}\orcidlink{0000-0003-1991-119X}}

\authorrunning{F.~Yu et al.}

\institute{Amazon \and Simon Fraser University \\}


{\maketitle
\centering
\includegraphics[width=0.99\textwidth]{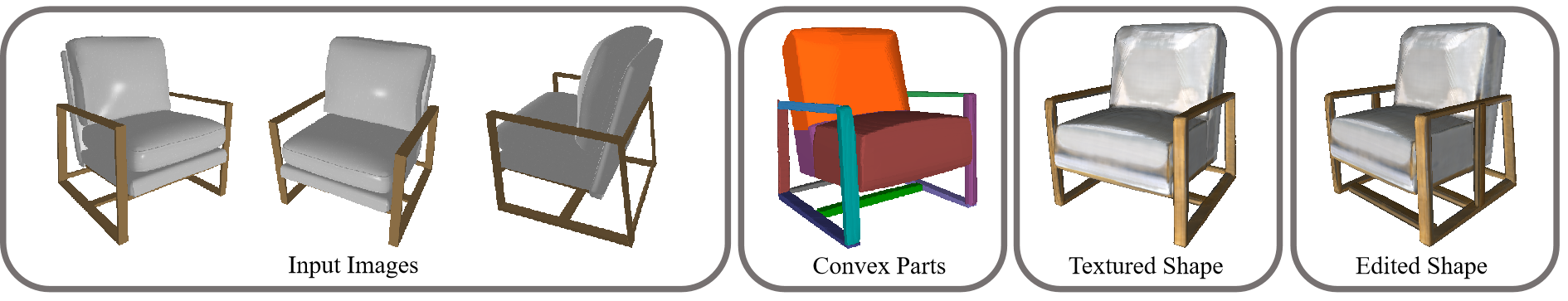}
\phantomsection
\captionof{figure}{Our method takes as few as three RGB images from disparate views and abstracts a textured 3D shape formed by a union of convex parts that well reflect the shape semantics. With a differentiable primitive assembly subject to only image-space losses, our network is trained without 3D supervision. With the meaningful parts abstracted, the resulting shape can be edited.}
\label{fig:teaser}

}
\begin{abstract}
We present a {\em differentiable rendering\/} framework to learn \rzz{{\em structured 3D abstractions\/} in the form of {\em primitive assemblies\/} from sparse} RGB images capturing a 3D object. By leveraging differentiable volume rendering, our method does not require 3D supervision. Architecturally, our network follows the general pipeline of an image-conditioned neural radiance field (NeRF) \rzz{exemplified by} pixelNeRF for color prediction. As our core contribution, we introduce {\em differential primitive assembly\/} (DPA) into NeRF to output a 3D occupancy field in place of density prediction, where the predicted occupancies serve as opacity values for volume rendering. Our network, coined DPA-Net, produces a union of convexes, each as an intersection of convex quadric primitives, to approximate the target 3D object, subject to an abstraction loss and a masking loss, both defined in the image space upon volume rendering. With test-time adaptation and additional sampling and loss designs aimed at improving the accuracy and compactness of the obtained assemblies, our method demonstrates superior performance over state-of-the-art alternatives for 3D primitive abstraction from sparse views.
\end{abstract}

\section{Introduction}
\label{sec:intro}

\rzz{3D reasoning, e.g., abstraction or reconstruction,} from single or multi-view images is one of the most fundamental problems in computer vision. With the recent
emergence of neural fields~\cite{NF_survey}, especially neural radiance fields (NeRF)~\cite{mildenhall2021nerf,muller2022instant} and
3D Gaussian splatting~\cite{kerbl2023gs}, rapid advances have been made in 3D reconstruction quality, speed, and the ability to take on sparse~\cite{yu2021pixelnerf,vora2023divinet,long2022sparseneus,yu2022monosdf} rather than
dense input views as in earlier works.
However, by design, NeRF and most of its variants target novel view synthesis with a focus on optimizing their primitives to improve
{\em rendering\/} performances, rather than serving downstream tasks involving shape {\em modeling\/} or {\em manipulation\/}.

Higher-level primitives than points and Gaussian splats are more suited to model shape structures~\cite{mitra_star13}, 
facilitating structure-level analysis and synthesis tasks such as semantic labeling, interactive shape editing, 
assembly-based modeling, and visual program induction~\cite{egstar2020_struct}.
Recently, several methods have been proposed for CAD modeling by learning primitive assemblies such as constructive solid geometry 
(CSG) trees~\cite{yu2022capri,yu2023d2csg,ren2021csgstump,kania2020ucsg}, sketch-n-extrude models~\cite{li2023secad}, or
shape programs~\cite{jones2022plad,ganeshan2023iccv}. However, these neural models all take 3D inputs such as voxels and point clouds.


In this paper, we introduce a learning framework which outputs \rzz{a {\em 3D abstraction\/}, in the form of a {\em primitive assembly\/},} from a small number of RGB images capturing a 3D object.
Importantly, the sparse input images are not only few in number, i.e., as few as {\em three\/}, but also {\em disparate\/}~\cite{vora2023divinet}, meaning that
the viewpoints may be significantly different; see Fig.~\ref{fig:teaser}. In addition to tackling this challenge, our method also does not require 3D supervision
for the multi-view 3D abstraction. It resorts to {\em differentiable rendering\/} of the assembled 3D primitives to define image-space abstraction losses.

\begin{figure*}[t!]
     \centering
     \includegraphics[width=0.99\linewidth]{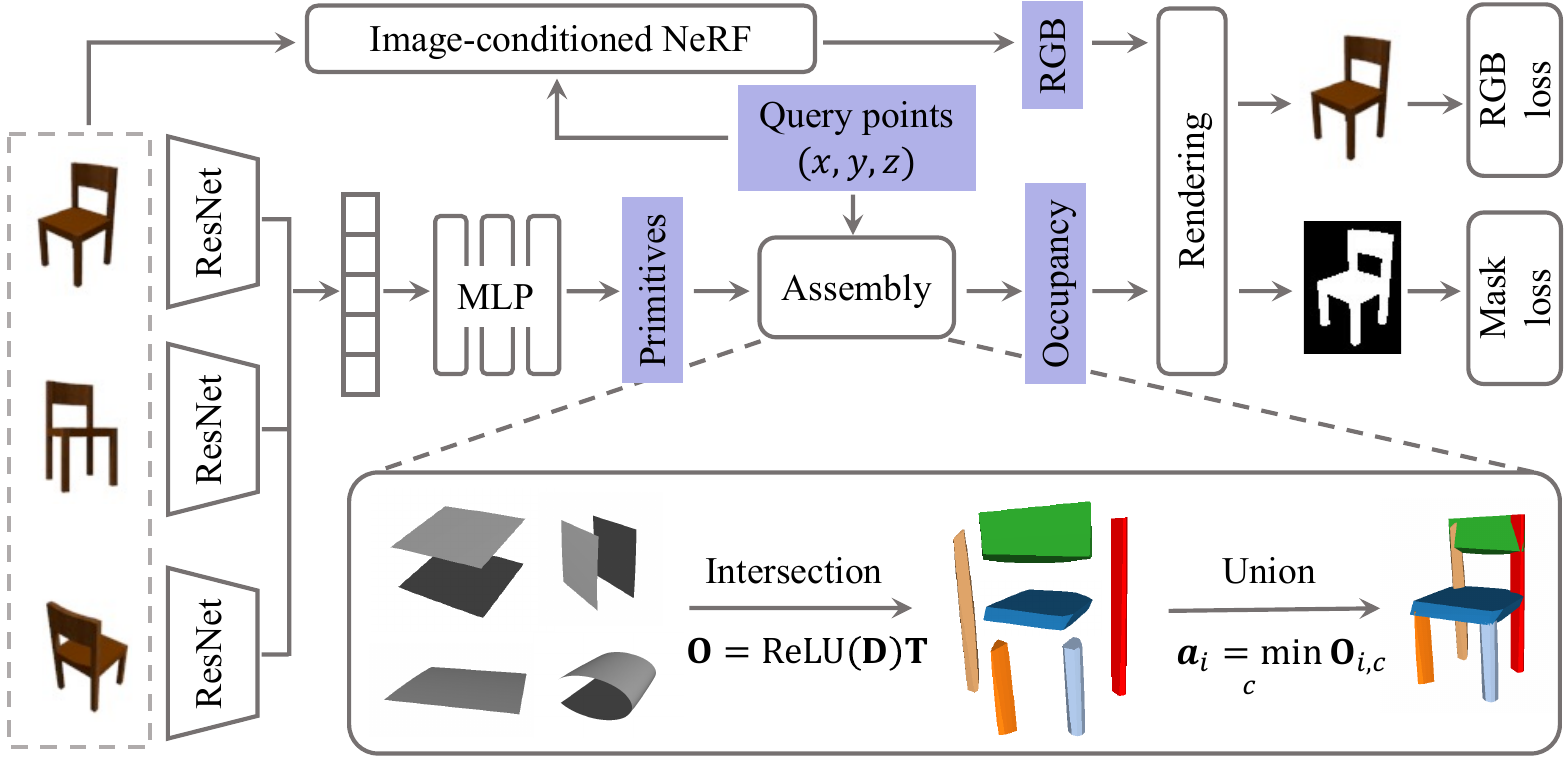}
   \caption{Overview of DPA-Net. Given a sparse set of input RGB images whose viewpoints can be significantly different, our network is trained to predict a 3D primitive assembly, \rzz{i.e., a 3D abstraction,} via differentiable volume rendering without 3D supervision. \rz{The high-level network architecture resembles that of an image-conditioned NeRF such as pixelNeRF~\cite{yu2021pixelnerf} for color prediction from multi-scale image features. What is new} is that the density estimation in NeRF is replaced by our novel differentiable primitive assembly (DPA). DPA takes as input multi-view image features from ResNet that are fused into a shape feature via \rz{weighted pooling}. The shape feature is further passed into an MLP (the primitive decoder) to predict the parameters of a set of convex quadric primitives. 3D query points and the primitives are assembled by two CSG- based assembly layers (intersection and then union) to predict point occupancies, which serve as opacity values for both volume rendering and for predicting an image mask. An RGB loss and a masking loss are calculated against the input images and object masks. Note that we assume that the camera poses and object masks are either provided or estimated prior to our 3D abstraction.
} 
\label{fig:overview}
\end{figure*}

Architecturally, our network follows the general pipeline of an image-conditioned NeRF (see Fig.~\ref{fig:overview}), such as 
pixelNeRF~\cite{yu2021pixelnerf}. In addition to 3D positions and camera ray information, we incorporate multi-scale image
features from a ResNet~\cite{he2016deep} encoder applied to a training set for differentiable rendering, enabling generalization 
across scenes.


Our new and core contribution is the introduction of {\em differential primitive assembly\/} (DPA) into the NeRF framework to output a {\em 3D occupancy field\/} 
in place of density prediction, so that the predicted occupancies would serve as opacity values for volume rendering. Specifically, our network, coined 
DPA-Net, produces a union of convexes, each as an intersection of convex quadric primitives, to approximate the target 3D object. The entire process requires no 
3D models for supervision; it takes fused multi-view image features as well as NeRF inputs to predict parameters for the CSG primitive assembly and 
the occupancy field, subject to an abstraction loss (named as RGB loss in Fig.~\ref{fig:overview}) and a masking loss, both defined in the image space upon volume rendering.

Furthermore, the abstraction of {\em clean} and {\em structurally\/} accurate primitive assemblies from sparse views offers several special 
challenges. To address them, we incorporate three enhancements: a) a primitive {\em dropout\/} scheme to improve compactness of the assembly;
b) an adaptive point sampling to assist in the recovery of thin structures; and c) an additional penalty to discourage primitive overlap. 

To assess our method, we train it on multi-view projections of 3D models from ShapeNet~\cite{chang2015shapenet} as well as real images from the DTU dataset~\cite{jensen2014large} for sparse-view abstraction.
We conduct qualitative and quantitative evaluations, ablation studies, as well as comparisons to state-of-the-art alternatives for 3D primitive abstraction from 
sparse-view images, demonstrating superior performance of DPA-Net. \rzz{The applications further demonstrate that users can easily edit our 3D abstraction results in popular CAD softwares, e.g., MeshLab~\cite{meshlab} and OpenSCAD~\cite{openscad}. The editable abstractions can serve as structural prompts and benefit other 3D generation tasks~\cite{chen2023shaddr,sella2023spic}.}

\section{Related Works}
\label{sec:related_work}

The literature on multi-view 3D \rzz{reasoning} and representation learning of 3D structures and abstractions is vast. In this section, we
focus on prior works that are closely related to our DPA approach, especially those designed to handle
sparse input views.


\subsection{Multi-view 3D reconstruction}

NeRF~\cite{mildenhall2021nerf}, most of its variants, and other multi-view neural 3D reconstruction methods have been designed for dense views~\cite{NF_survey}, 
but there are exceptions. SparseNeUS~\cite{long2022sparseneus} can take on input views that are few in number, \rzz{but {\em close-by\/},} whereas 
VolRecon~\cite{ren2023volrecon} improves the reconstruction quality with additional ground-truth (GT) depth maps as supervision. 
\rz{Fewer methods have been designed to handle sparse {\em and disparate\/} views.} In particular, FvOR~\cite{yang2022fvor} jointly solves for camera poses and 
3D geometry while requiring GT 3D shapes to supervise. FORGE~\cite{jiang2022few} improves FvOR by leveraging cross-view correlation, while using 2D images as 
supervision. Most recently, DiViNet~\cite{vora2023divinet} learns Gaussian surface priors, called neural templates, during training. At test times, it applies these 
templates as anchors to help stitch the surfaces over sparse regions. 
While FORGE and DiViNet both employ differentiable volume rendering as in our approach, their results are neither compact, in terms of primitive/Gaussian counts, nor 
well-structured to reflect shape semantics, as in the primitive assemblies obtained by our work.

As described above, our network architecture resembles that of pixelNeRF~\cite{yu2021pixelnerf}, which is \rz{an early and representative} method for generalizable
(i.e., not an overfit model) NeRF prediction from very few input views. \rz{Since then, there have been more recent works for sparse-view NeRFs, \eg, 
IBRNet~\cite{wang2021ibrnet}, RegNeRF~\cite{nie2022regnerf}, depth-supervised NeRF~\cite{deng2022dsnerf}, FreeNeRF~\cite{yang2023freenerf}, SPARF~\cite{truong2023sparf},
including those designed to handle single-view images, \eg, Pix2NeRF~\cite{cai2022pix2nerf} and Zero-1-to-3~\cite{liu20230123}.}
Comparing abstraction results from DPA-Net and those of pixelNeRF, we observe that their overall accuracies are comparable, but when the textures are removed, the mesh surfaces obtained from pixelNeRF results are quite noisy. \rz{It is important to note that our work is {\em not\/} intended to outperform all state-of-the-art methods on sparse-view 
3D reconstruction in terms of reconstruction quality. Foremost, we target neural 3D {\em primitive assembly/abstraction\/} while aiming to achieve a high reconstruction quality. For reference, we compare DPA-Net to both pixelNeRF and DiViNeT in Section~\ref{sec:exp}.}

\subsection{Learning 3D structures and abstractions}

The dominant majority of prior works on learning 3D shape structures~\cite{egstar2020_struct} take on 3D inputs, \eg, voxel grids, 
polygonal meshes, point clouds. This is also true for 3D abstractions designed for a variety of primitives such as cuboids~\cite{tulsiani2017cub}, 
super quadrics~\cite{pas2019sq,liu2022robust}, and convexes~\cite{chen2020bsp,deng2020cvx}. \rz{Among them, EMS (for Expectation, Maximization, and
Switching) is a probabilistic method to recover superquadrics from point clouds, which can be obtained from pure image inputs via multi-view stereo (MVS). Hence,
it is chosen as a baseline to evaluation our method in Section~\ref{sec:exp}.}

Most relevant to our approach are the series of works on unsupervised learning of CSG assemblies, including UCSG~\cite{kania2020ucsg}, 
CAPRI-Net~\cite{yu2022capri}, CSG-Stump~\cite{ren2021csgstump}, and most recently, D$^2$CSG~\cite{yu2023d2csg}, all of 
which optimize a CSG tree from 3D inputs. Our quadric-based primitive assembly is built on CAPRI-Net~\cite{yu2022capri}, with
our core contribution being to integrate it with differential volume rendering for DPA from few and disparate views.

Earlier works on structured CAD modeling or abstraction from images~\cite{xu2011photo,huang2015svr,izadinia2017im2cad} or sketches~\cite{xu2013s2s} 
all require 3D models to either form a repository for shape/part retrieval or serve as training data for supervised learning. Im2Struct~\cite{niu2018im2s} recovers 
3D shape structures, in the form of a cuboid assembly, from a single RGB image. Binary object masks were also used as an auxiliary signal to assist in
the structure recovery from input images, where the latter was trained with ground-truth 3D cuboid assemblies.
Also requires 3D models for training, StructureNet~\cite{mo2019structureNet} jointly embeds 3D cuboid representations and images into a common 
latent space to allow 3D cuboid abstractions from images.
With 3D supervision, BSP-Net~\cite{chen2020bsp} is able to predict a plane assembly from single-view RGB images. \rzz{Furthermore, RIM-Net~\cite{niu2022rim} and ~\cite{paschalidou2020learning} learn hierarchical part decompositions under 3D reconstruction supervision.}

To our knowledge, only three very recent works~\cite{monnier2023dbw,tertikas2023generating,alaniz2023iccv} have attempted 3D primitive abstraction from RGB images
without 3D supervision. \rz{PartNeRF~\cite{tertikas2023generating} learns a 3D primitive abstraction using ellipsoids, targeting 3D shape generation rather than reconstruction. Their use of ellipsoidal primitives also limits the method's ability to learn a meaningful part decomposition. For example, the top of a table is 
always split into multiple parts.}
ISCO~\cite{alaniz2023iccv} optimizes parameters for 3D superquadrics to recompose an object directly from 2D views, where the rendering loss is defined with respect to 2D 
image silhouettes. On few input views, \eg, four, their abstraction results were admittedly ``poor''; see Section 4.2 of their paper.
Since there is no published code by ISCO, we mainly compare our method with Differentiable Blocks World (DBW)~\cite{monnier2023dbw}, which can 
produce textured superquadric meshes. Overall, their results are compact, \ie, with few primitives, but the abstraction errors are relatively high.

\section{Approach}
\label{sec:method}

Given a small number of RGB images depicting a single object from disparate camera views, our goal is to learn a 3D shape abstraction assembled from a set of simple geometric primitives (quadrics in this paper). We assume that the camera parameters and object mask  for each view are known or pre-estimated. The cameras exhibit wide baselines and the number of images is typically less than six. 

Our method does not require \rz{any 3D supervision\/} such as GT primitive decompositions. \rz{The differentiable primitive assembly facilitates volume rendering and is trained with image-space losses only}, as shown in Fig.~\ref{fig:overview}.


\subsection{Feature extraction and aggregation}
\label{sec:feature}
For each input image, we employ ResNet34~\cite{he2016deep} as the feature encoder and alter its final linear layer to output a 256-dimensional feature vector. We then utilize the weighted pooling proposed in~\cite{wang2021ibrnet} to obtain an aggregated feature $\mathbf{f}\in\mathbb{R}^{256}$ from multi-view images. In practice, the weights are computed with a shared single-layer multi-layer-perceptron (MLP) followed by a softmax function.


\subsection{\rz{Primitive assembly}}
\label{sec:dpa}
Inspired by the state-of-the-art (SOTA) 3D primitive representation CAPRI-Net~\cite{yu2022capri}, we model a 3D shape as the assembly of a set of \emph{quadric} primitives using differentiable intersection and union operations. This step takes in the fused image feature $\mathbf{f}$ and a set of query points evenly sampled from camera rays as in NeRF~\cite{mildenhall2021nerf}, and predicts primitive parameters and an occupancy field of the shape---whether a query point is inside or outside the shape. It is noteworthy that our intersection and union operations are adopted from~\cite{yu2022capri,chen2020bsp}; we briefly discuss them here for completeness.

\mysubsubsection{Primitive parameterization} We define a quadric primitive as: $|a|x^2+|b|y^2+|c|z^2+dx+ey+fz+g=0$, 
where $(x,y,z)$ is a query point and $(a,b,c,d,e,f,g)$ are the primitive parameters. We enforce the first three parameters to be non-negative to include convex quadric primitives only. Denote the number of primitives as $P$, we obtain a matrix $\mathbf{P}\in\mathbb{R}^{P\times7}$ by stacking all primitive parameters. Then a 2-layer MLP is applied to infer $\mathbf{P}$ from the fused feature $\mathbf{f}$. We set $P=4,096$ in this paper and please refer to the supplementary for an ablation study.

\rz{Define matrix $\mathbf{Q}$ whose $i$th row 
$(x^2_i,y^2_i,z^2_i,x_i,y_i,z_i,1)$ is}
computed from the $i$th query point $(x_i,y_i,z_i)$. \rz{With $N$ query points}, we estimate an $N\times P$ matrix $\mathbf{D}=\mathbf{Q}\mathbf{P}^\top$, where $\mathbf{D}_{i,p}$ denotes the approximate signed distance between the $i$th point and the $p$th primitive. We have $\mathbf{D}_{i,p}<0$ if point \rz{$i$} is \emph{inside} the primitive and $\mathbf{D}_{i,p}>0$ if \emph{outside}.

\mysubsubsection{Primitive intersection} We adaptively group the $P$ quadric primitives into $C=256$ convexes. 
Our goal is to check whether a query point is inside any convex. To this end, we utilize a \emph{learnable} selection matrix $\mathbf{T}\in\mathbb{R}^{P\times C}$. Ideally, $\mathbf{T}$ is binary, but may take floating-point values in $[0,1]$ depending on the training phase; see Sec.~\ref{sec:training}.  
Each column in $\mathbf{T}$ selects a subset of primitives whose intersection forms one of the $C$ convexes. 
In practice, we calculate an $N\times C$ matrix $\mathbf{O}=\text{ReLU}(\mathbf{D})\mathbf{T}$, with the ReLU function. We set $\mathbf{O}_{i,c}=0$ if the $i$th point is inside the $c$th convex, and $\mathbf{O}_{i,c}>0$ if outside.


\mysubsubsection{Union of convexes} We further group the convexes to form the final 3D shape, checking whether a query point is inside this shape (\ie, the occupancies). We attain this with a min-pooling among all convexes for the $i$th point,
\rz{analogous to the union in CSG:}
\begin{equation}
    \mathbf{a}^*_i = \min_{1 \leq c \leq C} \mathbf{O}_{i,c}
    \hspace{0.35cm}
    \begin{cases}
    = 0 & \text{inside,} \\
    > 0 & \text{outside,}
    \end{cases}
    \label{eq:label_a_*}
\end{equation}
where $\mathbf{a}^*\in\mathbb{R}^N$ denotes the occupancy field. 
In practice, 
a soft approximation is calculated for better gradient flow:
\begin{equation}
    \mathbf{a}^+_i = \mathscr{L}\left(\sum_{c=1}^C \mathbf{w}_c  \mathscr{L}\left(1 - \mathbf{O}_{i,c}\right)\right)
    \hspace{0.1cm}
    \begin{cases}
    =1 & \approx  \text{inside,} \\
    <1 & \approx \text{outside,}
    \end{cases}
    \label{eq:label_a_+}
\end{equation}
where $\mathbf{w} \in \mathbb{R}^{C}$ is a learnable weight vector with elements close to $1$ and $\mathscr{L}()$ a function that clips its input to the range $[0, 1]$. Notice that $\mathbf{a}^+_i$ also corresponds to the opacity value in volume rendering~\cite{mildenhall2021nerf,ren2023volrecon}. Refer to the supplementary for examples of the learned primitives and convex shapes.

\subsection{Differentiable rendering}
\label{sec:color}
We follow pixelNeRF \cite{yu2021pixelnerf} to compute the color for each query point. We use the same network specification, except \pg{we remove} the volume density prediction branch. Specifically, given a pixel in \rz{an input view}, we obtain the RGB color $\widehat{\mathbf{c}}$ of each query point along the corresponding ray using MLPs. The inputs include the 3D position, ray direction, and image features projected from all views. 


By accumulating \rz{over} all query points \rz{$i$} along that ray, we render the color at the given pixel as:
$\widehat{\mathbf{C}} = \sum_{i=1}^{R}t_i\alpha_i\widehat{\mathbf{c}}_i,$
where $R$ is the number of sampled points along \rz{the} ray, 
$\alpha_i$ denotes the opacity which takes on different values at different phases of the optimization (see Sec.~\ref{sec:training}), and $t_i=\prod^{i-1}_{k=1}(1 - \alpha_k)$ is the accumulated transmittance. We also render the object mask value as $\widehat{\mathbf{M}} = \sum_{i=1}^\Omega t_i\alpha_i$.

\subsection{Network training and test-time adaptation}
\label{sec:training}
Similar to~\cite{yu2021pixelnerf,long2022sparseneus,ren2023volrecon}, DPA-Net can work in a \emph{feed-forward} fashion by training on a dataset including sparse-view images. We also propose a multi-phase test-time adaptation for instance-specific refinement during inference.

\mysubsubsection{Pretraining} Our loss function consists of three terms:
\begin{align}
    \label{eq:loss_all}
    \mathcal{L} & = \mathcal{L}_{ph} + \mathcal{L}_{\mathbf{T}} + \mathcal{L}_{\mathbf{w}}, \\ 
    \mathcal{L}_{ph} & =  \frac{1}{B} \sum_{j=1}^{B} \left\| \widehat{\mathbf{C}}_j - \mathbf{C}_j \right\|_2^2 + \frac{1}{B} \sum_{j=1}^{B} \left\| \widehat{\mathbf{M}}_j - \mathbf{M}_j \right\|_2^2, \\
    \mathcal{L}_{\nv{T}} & = \sum_{i,j} \text{max}(-\mathbf{T}_{i,j}, 0) + \text{max}(\mathbf{T}_{i,j} -1, 0), \\
    \mathcal{L}_{\nv{w}} & = \sum_{j} |\nv{w}_j -1|,
\end{align}
where $\mathcal{L}_{ph}$ including an abstraction loss and a mask loss enforces photo consistency, $B$ is \#pixels in a training batch, and $\nv{C}$ and $\nv{M}$ denote GT color and mask, respectively. We define the opacity $\alpha=\mathbf{a}^+$ from Eq.(\ref{eq:label_a_+}). $\mathcal{L}_{\nv{T}}$ keeps entries of the selection matrix $\nv{T}$ within range $[0,1]$. $\mathcal{L}_{\nv{w}}$ ensures that entries of the weight vector $\nv{w}$ close to $1$. All training instances share the same $\nv{T}$ and $\nv{w}$. 

\mysubsubsection{Test-time adaptation (TTA)} We have found that DPA-Net trained on a generic dataset risks suboptimal performance when applied to novel shapes which deviate substantially from the training data. For example, assembling each 3D object requires unique matrices 
$\nv{T}$ and $\nv{w}$ tailored to that instance. To improve generalization, we propose a test-time adaptation that operates in three phases with different configurations as shown in Table~\ref{tab:tta}. Each phase is initialized from its preceding phase.

\noindent \underline{Phase 1:} The same as pretraining except that the input test images are used as GT.

\noindent \underline{Phase 2:} Instead of using the approximated occupancy field $\mathbf{a}^+$, we opt to \rz{employ} the exact definition $\mathbf{a}^*$ to further improve the abstraction performance. The loss function becomes $\mathcal{L}_{ph} + \mathcal{L}_{\mathbf{T}}$, whereas $\mathcal{L}_{\mathbf{w}}$ is omitted since $\mathbf{a}^*$ is devoid of $\mathbf{w}$ as in Eq.(\ref{eq:label_a_*}). We set the opacity $\alpha=\exp{(-10\nv{a}^*)}$.

\noindent \underline{Phase 3:} Ideally, the selection matrix $\nv{T}$ should be binary to exactly mimic the operation of quadric primitive intersection. 
To ensure that, we follow~\cite{chen2020bsp,yu2022capri} to binarize $\nv{T}$ with a threshold of $0.01$ and freeze it at this phase. We then fine-tune the network with two loss terms: the photo-consistency term $\mathcal{L}_{ph}$ and an additional regularization term $\mathcal{L}_{over}$ that discourages excessive overlap between convex shapes. 

\rz{\mysubsubsection{Overlapping loss}}
As observed in~\cite{monnier2023dbw,paschalidou2021neural}, it is non-trivial to explicitly calculate and penalize \rz{area overlaps. Instead, we} constrain points lying inside overlapped areas. Concretely, we define $\mathbf{h}_i=\sum_{c=1}^C \exp{(-10\mathbf{O}_{i,c})}$ for the $i$th point, where $\mathbf{h}_i$ is large if the point lies within multiple convexes. We minimize $\mathcal{L}_{over} = \frac{1}{\#\Omega} \sum_{i\in\Omega}\text{max}(\mathbf{h}_i, 1.9)$, where $\Omega$ is the set of indexes of 3D points inside the shape, with cardinality $\#\Omega$. In practice, we sample 40,960 points surrounding the recovered shape in Phase 2 and utilize the condition $\mathbf{a}_i^*<0.01$ to check point membership in $\Omega$.

\begin{table*}
\centering
\caption{Configurations for multi-phase test-time adaption (TTA).}
\begin{tabular}{c c c c c c c c}
\toprule
Phase & Type of $\nv{T}$ & Type of $\nv{w}$ & Occupancy & Opacity & Dropout & Loss  & Model parameters\\
\midrule
1 & float  & float  & $\nv{a}^+$ & $\nv{a}^+$ & - & $ \mathcal{L}_{ph} + \mathcal{L}_{\nv{T}} + \mathcal{L}_{\nv{w}}$ & network, $\nv{T}$, $\nv{w}$ \\
2 & float  & -      & $\nv{a}^*$ & $\text{exp}(-10\nv{a}^*)$ & - & $\mathcal{L}_{ph} + \mathcal{L}_{\nv{T}}$ & network, $\nv{T}$ \\
3 & binary & - & $\nv{a}^*$ & $\exp{(-10\nv{a}^*)}$ & \checkmark & $\mathcal{L}_{ph} + \mathcal{L}_{over}$ & network \\
\bottomrule
\end{tabular}
\label{tab:tta}
\end{table*}

\section{\rz{Network Improvements}}
\label{sec:improments}

\rz{The core modules of DPA-Net as described in Sec.~\ref{sec:method} can still underperform when abstracting objects with complex structures. One issue comes from objects with holes and the other is redundant primitives.
To address these issues, we present two key improvements to our network.}

\mysubsubsection{Silhouette-aware pixel sampling} 
\rz{By default, NeRF~\cite{mildenhall2021nerf} and variants randomly sample pixels during optimization, \rzz{which was to ensure a balanced learning. However,} we found DPA-Net to struggle with it when abstracting 3D objects with complex topologies, \eg, thin structures or holes.} To address this, we adopt adaptive pixel sampling \rzz{to allocate more pixels near} {\em object silhouettes\/}. Specifically, we sample two pixel sets: 256 pixels randomly across the entire raster space, and 1,000 pixels around object contours. The contour pixels are obtained by performing morphological closing on the mask image and then adding Gaussian noise to the pixel coordinates.

\rz{Such an adaptive sampling scheme places more emphasis on boundary areas during optimization to improved abstraction amid complex 3D topologies. We believe that it is a {\em general\/} sampling strategy which could potentially benefit other abstraction frameworks other than DPA-Net. We leave this investigation to future work.}




\mysubsubsection{Primitive dropouts}
\rz{Attaining a compact primitive assembly, with a low primitive count, without compromising abstraction accuracy is clearly desirable.}

Similar to~\cite{yu2023d2csg}, we drop out a primitive if excluding it from the construction does not significantly alter the overall shape. In practice, we initialize with the primitives obtained from Phase 3. For each $p$th primitive, we set the $p$th row of the binary selection matrix $\mathbf{T}$ to zero, compute a new occupancy field $\mathbf{b}^*$ using Eq.\ref{eq:label_a_*}, and finally quantify the shape variation as:
    $v = \frac{1}{K}\sum_{i=1}^{K} \mathbbm{1}\left(\text{XOR}(\mathbf{b}_i^*<0.1, \mathbf{a}_i^*<0.1)\right),$
where $K=40,960$ is the number of points sampled around the shape as in Phase 3 and $\mathbbm{1}()$ maps Boolean values to \rz{0 or 1.} 

If $v$ is close to zero, we keep the $p$th row at zero. Otherwise, we restore the original values. We iterate through each row of \rz{$\mathbf{T}$} to retain the set of primitives corresponding to non-zero rows, where this primitive dropout is applied at every 400 iterations during Phase-3 fine-tuning. Our experiments show that this strategy can remove up to 29\% redundant primitives; see supplementary for more details.

\section{Results \& Evaluation}
\label{sec:exp}

We implement DPA-Net in PyTorch~\cite{pytorch}, using the Adam~\cite{adam} optimizer with a learning rate of 0.0001. Each batch contains 4 object instances. Our model is trained and evaluated on 4 NVIDIA Tesla V100 GPUs. We evaluate on \kai{synthetic images rendered from} ShapeNet~\cite{chang2015shapenet} and real \kai{images from the} MVS DTU benchmark~\cite{jensen2014large}, \kai{using}
\ym{standard metrics for 3D abstraction quality and shape compactness.} Pre-training on ShapeNet takes about three days. TTA on average takes one hour for ShapeNet and two hours for DTU. 


We compare our method with \rzz{two recent} 3D shape abstraction methods, \rzz{EMS~\cite{liu2022robust} and DBW~\cite{monnier2023dbw}}, for which open-source code is available\footnote{Note that the code of ISCO~\cite{alaniz2023iccv} was not publicly available upon submission.}.
In addition, we also compare with \rzz{two reconstruction methods, pixelNeRF~\cite{yu2021pixelnerf} and DiViNet~\cite{vora2023divinet}.}

\noindent {\bf EMS}~\cite{liu2022robust} is an optimization-based method for fitting super-quadrics on 3D points. We sample 6K points from the GT meshes and use parameter settings from~\cite{liu2022robust}. 

\noindent {\bf DBW}~\cite{monnier2023dbw} is the SOTA method to recover super-quadric meshes from \emph{dense-view} images. We adapted it for sparse views but failed to obtain meaningful results. 
For a fair comparison, we instead compare to it using intended dense image inputs on DTU.

\noindent {\bf pixelNeRF}~\cite{yu2021pixelnerf} is an image-conditioned NeRF for novel view synthesis from sparse images. In its original implementation, the volume density is predicted as a function of both location and viewing direction. However, the density value for each point should be independent of the viewing direction~\cite{mildenhall2021nerf}, which motivates us to modify its network architecture to learn better volume density field for subsequent geometry extraction. In particular, we predict the density using the first three ResBlocks that do not take the viewing direction as input. We then extract the final mesh by applying marching cubes on the inferred density field. Note that the output geometry is un-structured.

\noindent {\bf DiViNet}~\cite{vora2023divinet} \rzz{is the SOTA method for neural 3D surface reconstruction from sparse- {\em and\/} disparate-view RGB images, via volume rendering. It uses learned 3D Gaussians as shape template priors and ``stitch" the shape surface to the template.}

\mysubsubsection{Metrics}
We evaluate our method from \fg{three} aspects. 
(1) \kai{3D abstraction quality, measured by three standard metrics}~\cite{chen2020bsp,yu2022capri}: symmetric Chamfer Distance (CD), Edge Chamfer Distance (ECD), and Normal Consistency (NC). The same parameter settings from CAPRI-Net~\cite{yu2022capri} are used to calculate CD and ECD. 
(2) 
\kai{Shape compactness, \rzz{which is reflected by having fewer parts (\#Parts) and leads to ease of manipulation/editing.}} \fg{(3) Rendering quality, measured by the standard image quality metrics PSNR and SSIM.}

\subsection{Evaluation on ShapeNet}

We first evaluate on 
ShapeNet~\cite{chang2015shapenet} with
pre-rendered ShapeNet images from SRN~\cite{sitzmann2019srns} and Kato \etal~\cite{kato2018neural}. We use GT camera poses for both training and testing, as well as object masks from pixelNeRF, which were obtained via thresholding.

\mysubsubsection{Category-specific setting}
In this setting, we evaluate the methods on a single chair category.
We follow pixelNeRF and adopt the training/testing split used in SRN, with a total of 6,591 chairs. Each chair model is rendered from 50 different viewpoints to generate the training images. 
\kai{At training time, we randomly samples 3 views for each training step.}
At test time, we use 2 front-view images and 1 back-view image for each test chair. All images have $128\times 128$ resolution.
We compare DPA-Net against EMS~\cite{liu2022robust} and pixelNeRF~\cite{yu2021pixelnerf}. 
Table~\ref{tab:srn_chair} presents the main quantitative results.
EMS~\cite{liu2022robust}, \kai{while being a} state-of-the-art primitive fitting methods from 3D inputs, produces larger errors across all metrics, while using 
a significantly larger number of 3D parts. 

In particular, DPA-Net is better by 88\%/17\%/68\% in CD/NC/ECD and uses 45\% fewer 3D parts. 
Compared with the modified pixelNeRF, we obtain similar errors (differs by 0.3 in CD, 0.07 in NC, 0.8 in ECD, \fg{0.55 in PSNR and 0.003 in SSIM}) across all metrics. However, our method is more flexible in that it also produces convex part decomposition in an unsupervised manner, which is more useful in many applications, compared to the non-structured mesh produced by pixelNeRF.

\begin{figure}[t!]
     \centering
     \includegraphics[width=0.99\linewidth]{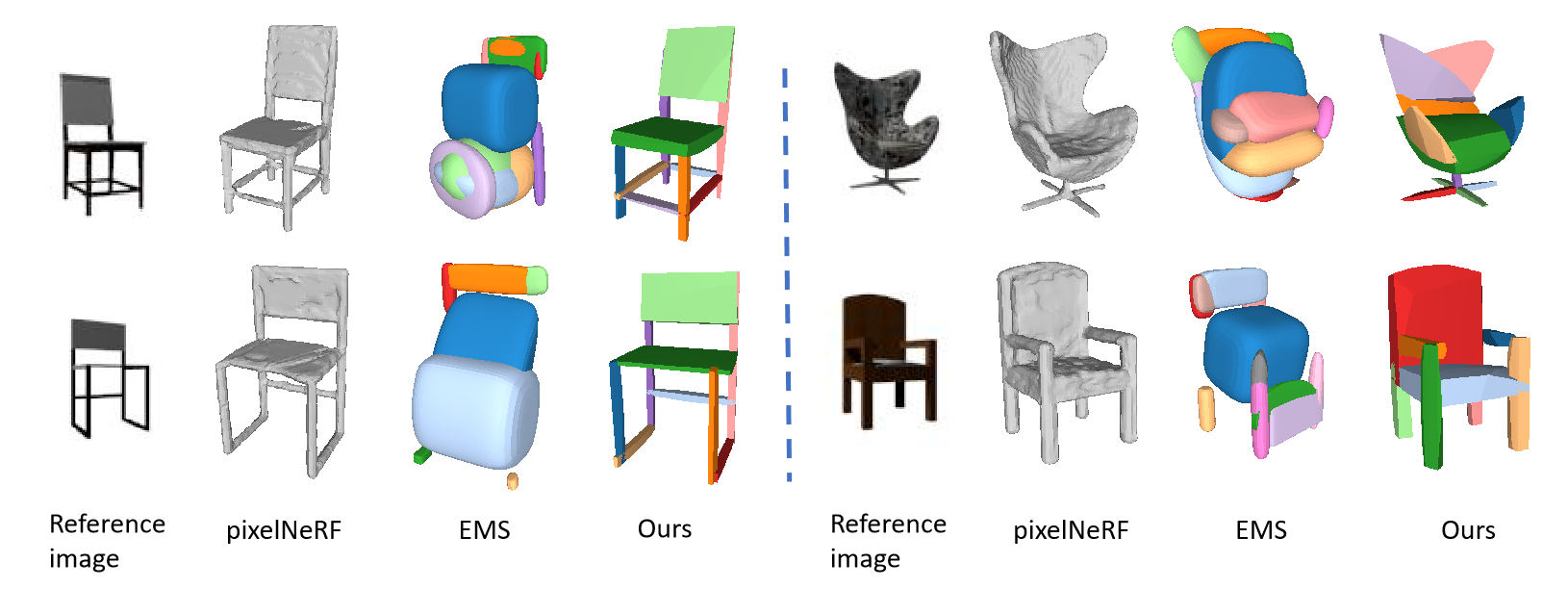}
   \caption{Visual comparisons on the ShapeNet chair benchmark. The \textcolor{red}{red} ovals highlights noisy surfaces, which can be seen more clearly if the image is zoomed in upon.}
\label{fig:chair_srn}
\end{figure}

The qualitative results in Fig.~\ref{fig:chair_srn} consolidates our conclusion above. Both pixelNeRF and DPA-Net are able to accurately recover the object geometry. However, due to the non-structured 3D representation, pixelNeRF suffers from noisy surfaces (highlighted by the \textcolor{red}{red} ovals). 
EMS is able to create smooth surfaces but exhibits inferior 3D abstraction quality. In contrast, DPA-Net not only abstracts reasonably accurate 3D shape but also produces clean and meaningful shape structure decomposition.

\begin{table*}[t!]
\centering
\caption{Quantitative comparisons against EMS~\cite{liu2022robust} and pixelNeRF~\cite{yu2021pixelnerf} on the ShapeNet chair benchmark. EMS takes 3D point cloud as input while pixelNeRF and our DPA-Net take 2D images. Since pixelNeRF outputs non-structured shapes, we mark with a ``-" in the \#Parts entry. The colors \first{cyan} and \second{orange} represent the best and the second best results. }
\begin{tabular}{l|c|c|c|c|c|c|c} \hline
Method & Input & CD $\!\downarrow$ & NC $\!\uparrow$ & ECD $\!\downarrow$ & \#Parts $\!\downarrow$ & PSNR $\!\uparrow$ & SSIM $\!\uparrow$ \\  \hline \hline
EMS & 3D & 6.93 & 0.65& 12.56 & 14.92 & - & - \\ \hline 
pixelNeRF & \multirow{2}{*}{2D} & \first{0.50} & \first{0.85} & \first{3.21} & - & \first{27.85} & \first{0.963} \\
Ours & & \second{0.79} & \second{0.78} & \second{4.01} & \first{8.15} & \second{27.30} & \second{0.960} \\ \hline
\end{tabular}
\label{tab:srn_chair}
\end{table*}



\mysubsubsection{Ablation studies}
We validate the effectiveness of \fg{four} important strategies on improving DPA-Net: test-time adaptation (TTA), adaptive pixel sampling (AS), overlapping loss (OL) and primitive dropout (PD). 
Table~\ref{tab:ablation} shows the main quantitative results. \fg{Removing TTA (w/o TTA) and adaptive sampling (w/o AS) significantly degrades the abstraction quality as the CD increases by 54\% and 41\% respectively, validating the contribution of our TTA strategy and silhouette-aware pixel sampling method.}
Removing the overlapping loss (w/o OL) results in higher number of primitives and convex shapes, showing that this loss helps in removing redundant, heavily overlapped convexes. The primitive dropout (PD) strategy further reduces the number of primitives by 29\% without sacrificing the abstraction accuracy.

\begin{figure}[t!]
     \centering
     \includegraphics[width=0.99\linewidth]{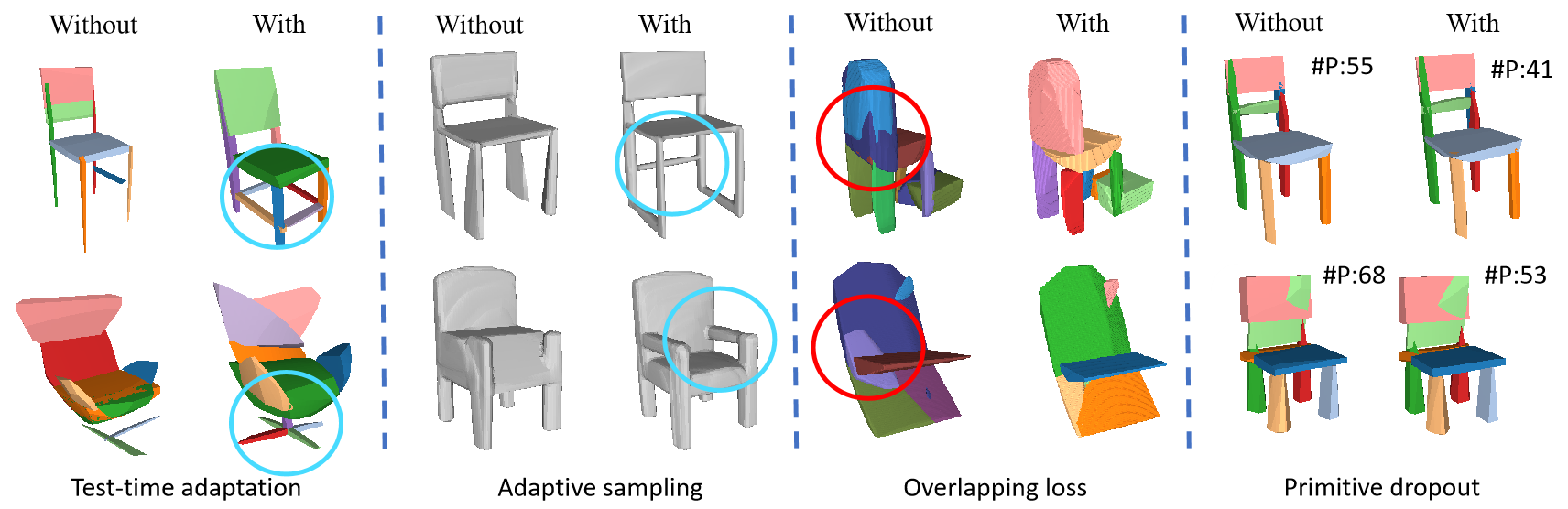}
   \caption{Ablation studies on the three key strategies to boost DPA-Net. \rzz{The texts, \#P's, in the last column} show the number of qudaric primitives used. \fg{The \textcolor{cyan}{cyan} ovals highlight challenging reconstruction areas that our method with TTA or adaptive sampling correctly recovers. The \textcolor{red}{red} ovals highlight redundant overlapping convex parts.}}
\label{fig:ablation}
\end{figure}

Fig.~\ref{fig:ablation} qualitatively demonstrates the contributions of the four designs. \fg{TTA helps recover missing structures from the pretraining results, while adaptive pixel sampling aids in capturing thin structures and achieving accurate shape silhouettes.} Enforcing a overlap loss removes redundant 3D parts. Primitive dropouts effectively reduce the number of primitives for assembling the final shape.

\begin{table*}[t!]
\centering 
\caption{Ablation studies on the ShapeNet chair benchmark. \#P and \#C denote the number of quadric primitives and convex shapes used during shape assembly. The colors \first{cyan} and \second{orange} represent the best and the second best results.}

\begin{tabular}{l|c|c|c|c|c} \hline 
Method & CD $\downarrow$ & NC $\uparrow$ & ECD $\downarrow$ & \#P $\downarrow$ & \#C $\downarrow$ \\ \hline\hline
w/o TTA & 1.22 & \second{0.81} & 6.11 & \first{39.04} & \first{7.38}\\
w/o AS & 1.11 & 0.75 & 5.31 & 45.81 & 8.23\\
w/o OL & 0.83 & \first{0.82} & 4.03 & 61.23 & 8.61\\
w/o PD & \second{0.81} & 0.78 & \first{3.85} & 59.64 & 8.45\\ 
Full model & \first{0.79} & 0.78 & \second{4.01} & \second{42.38} & \second{8.15}\\ \hline 
\end{tabular}

\label{tab:ablation} 
\end{table*}



\mysubsubsection{Category-agnostic setting}
For experiments in a {\em category-agnostic\/} way, we train our network on 13 largest categories of ShapeNet, covering roughly 43K object instances. We again follow pixelNeRF and adapt the training/testing split in Kato \etal~\cite{kato2018neural}, in which multi-view images, at resolution $64\times 64$, are rendered from 24 fixed elevated angles. During training, we randomly pick 3 views from those 24 views for each object.

Table~\ref{tab:shapenet13} summarizes the main results.
DPA-Net consistently outperforms EMS across all metrics by a significant margin, \rzz{except for NC}, reinforcing our category-specific findings. \rzz{DPA-Net being beat on NC is primarily due to the challenges posed by sparse view supervision. For instance, the table top in our results is not perfectly flat, which is a consequence of fine-tuning with a limited number of views. In contrast, EMS, which utilizes points from GT meshes, are more apt at fitting well-oriented primitives.}

More importantly, while pixelNeRF attained higher accuracy when trained solely on chairs, our DPA-Net surpasses pixelNeRF across all metrics (except for NC and PSNR) in this diverse multi-category experiment --- for instance, improving on the CD metric by nearly 60\%. This highlights the enhanced capacity of our structured primitive-based representation in capturing expansive shape variations compared to pixelNeRF's unstructured approach. 

\begin{figure}[t!]
     \centering
     \includegraphics[width=0.99\linewidth]{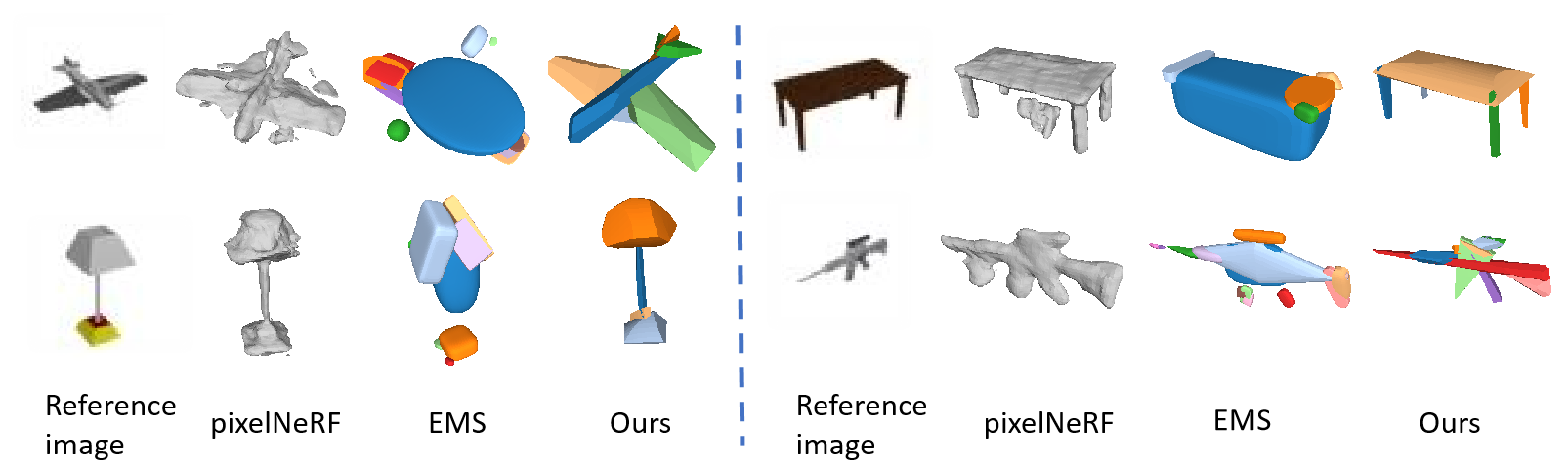}
   \caption{Qualitative comparisons in the category-agnostic setting. The image resolution is $64\times 64$.}
\label{fig:shapenet}
\end{figure}

As shown in Fig.~\ref{fig:shapenet}, DPA-Net successfully recovers 3D primitive assemblies closely approximating GT shapes despite relatively low $64\times64$ input image resolution. In summary, our multi-category evaluations further validate the effectiveness DPA-Net in generalized shape abstraction and structured primitive discovery from limited visual data across substantially diverse object classes.


\begin{table*}[t!]
\centering
\caption{Quantitative evaluations in the category-agnostic setting on the ShapeNet dataset. Results marked by the color \first{cyan} are the best.}

\begin{tabular}{l|c|c|c|c|c|c|c} \hline
Method & Input & CD $\!\downarrow$ & NC $\!\uparrow$ & ECD $\!\downarrow$ & \#Parts $\!\downarrow$ & PSNR $\!\uparrow$ & SSIM $\!\uparrow$ \\  \hline \hline
EMS & 3D & 5.15 & \first{0.75} & 43.45 & 8.28 & - & - \\ \hline 
pixelNeRF & 2D & 3.67 & 0.73 & 11.01 & - & \first{27.17} & 0.923 \\ \hline
Ours & 2D & \first{1.47} & 0.70 & \first{4.91} & \first{5.57} & 25.96 & \first{0.932} \\ \hline
\end{tabular}

\label{tab:shapenet13}
\end{table*}


  

\subsection{Evaluation on DTU}

To further demonstrate the applicability of our method for learning primitive assemblies, we also test on the real images from the DTU dataset~\cite{jensen2014large}, which contains multi-view scene images with curated backgrounds captured in a controlled indoor environment. 
Each scene contains one or more objects. 
We use the camera poses and object masks from DVR~\cite{niemeyer2020differentiable} and
resize all images to $300\times 400$. 
Following standard practices, we treat the structured light scans as 3D GT. 
\kai{Instead of pre-training on DTU, we apply TTA directly on the 6 scenes with object masks used in DBW~\cite{monnier2023dbw}.}
\kai{This challenging setting allows us to rigorously test the generalizability of DPA-Net to real images.}

\begin{figure}[t!]
     \centering
     \includegraphics[width=0.99\linewidth]{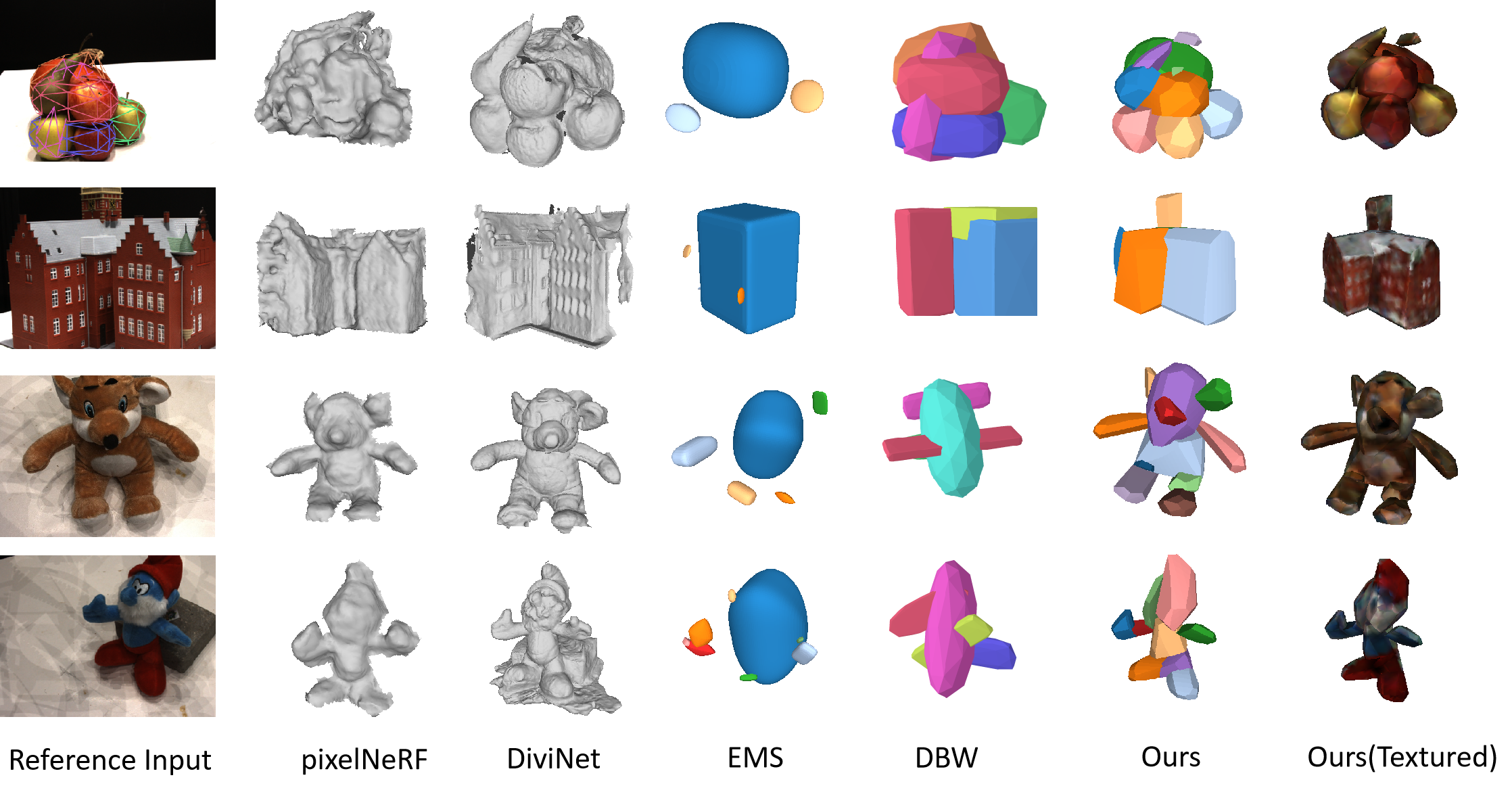}
   \caption{Visual comparisons on 3D abstraction and reconstruction over the DTU dataset.}
\label{fig:dtu}
\end{figure}

As shown in Table~\ref{tab:dtu}, DPA-Net obtains higher-quality 3D abstractions with 18\% lower average CD while using 3.5 more parts compared to DBW. DiViNet~\cite{vora2023divinet} and pixelNeRF~\cite{yu2021pixelnerf} outperform DAP-Net as they output un-structured meshes. Fig.~\ref{fig:dtu} provides visual comparisons showing that our method can recover fine structural details.


\begin{table}\centering
\caption{Quantitative comparisons on 6 DTU scenes with object masks. As DiViNet~\cite{vora2023divinet} outputs un-structured shapes, we mark with a ``-" in the \#Parts entry. \first{Cyan} marks the best results.}
 \begin{tabular}{l c c c c c c c c c c}
\toprule
Method & Input & Structure & S24 & S40 & S55 & S63 & S83 & S105 & Mean CD & Mean \#Parts \\ \midrule
EMS & 3D & \checkmark & 4.04 & 5.85 & 4.13 & 4.99 & 3.54 & 5.99 & 4.75 & 4.50\\
DBW & 2D & \checkmark &  \first{3.85} &	 \first{1.34} & 3.83	& 4.57 & 5.05 & 6.07 & 4.12 & \first{3.83} \\
Ours & 2D & \checkmark & 4.74 & 3.27 &  \first{2.88} &  \first{2.99} & \first{3.26} & \first{3.30} & \first{3.41} & 7.33 \\ 
\midrule
pixelNeRF & 2D & $\times$ & 3.54 & 2.07 & 2.27 & 4.57 & 3.52 & 2.12 & 3.01 & - \\
DiViNet & 2D & $\times$ & 3.34 & 1.47 & 0.75 & 2.74 & 1.94 & 1.17 & 1.90 & - \\

\bottomrule
\end{tabular}
\label{tab:dtu}
\end{table}

\subsection{Applications}

\begin{figure}[t!]
     \centering
     \includegraphics[width=0.9\linewidth]{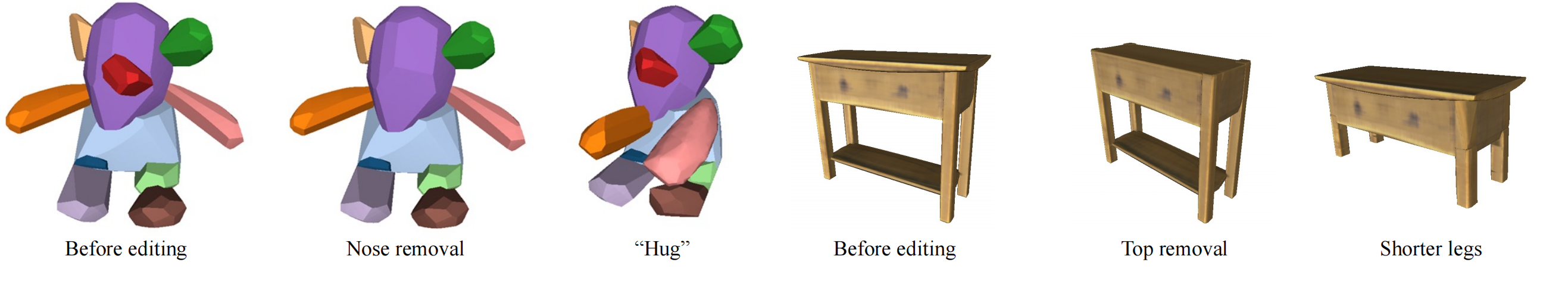}
   \caption{\rzz{Direct editing, using MeshLab~\cite{meshlab}, over structured abstractions obtained by DPA-Net.}}
\label{fig:edit}
\end{figure}

\begin{figure}[t!]
     \centering
     \includegraphics[width=0.9\linewidth]{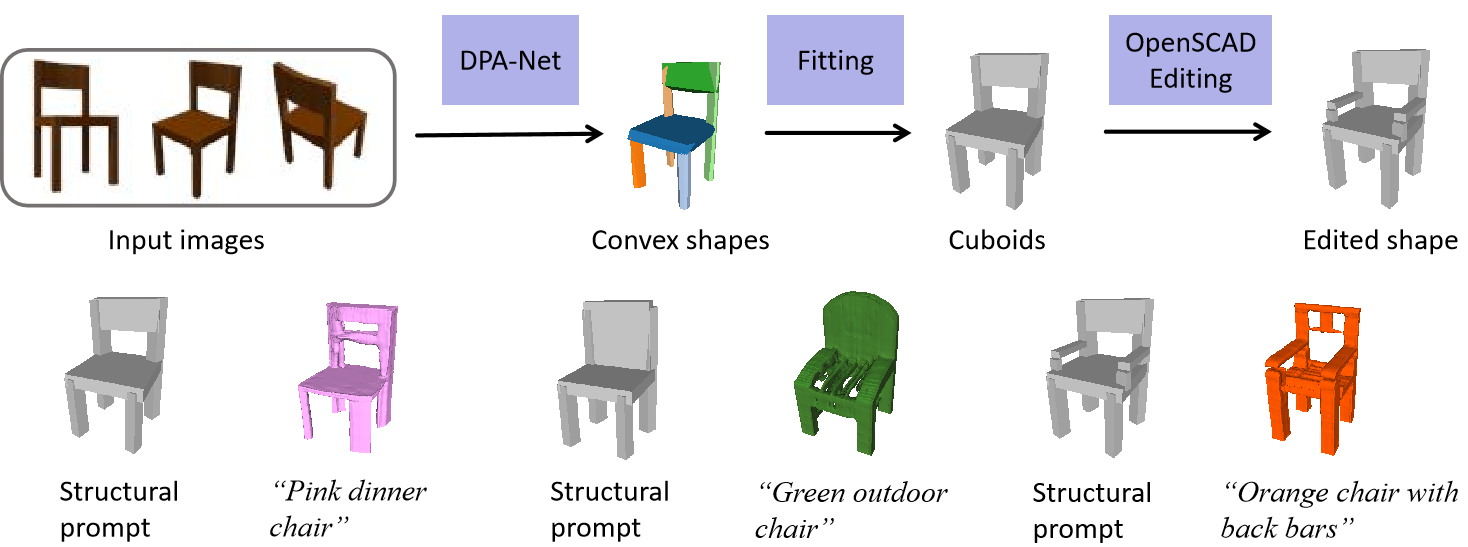}
   \caption{Top: editing 3D shape from DPA-Net by OpenSCAD~\cite{openscad}. Bottom: 3D shape generation by SPiC-E~\cite{sella2023spic} from text prompts and 3D guidance shapes as structural prompts. The generated detailed shapes share a similar structure as the input 3D guidance shapes and have more details according to the text condition.}
\label{fig:generation}
\end{figure}

\rzz{The structured 3D abstractions obtained by DPA-Net are fully interpretable since they represent primitives assembled by meaningful CAD operations. This allows the obtained results to be directly edited or animated for down-streaming applications. Users can easily modify the abstracted parts by adding, removing, translating, or rotating components to create desired CAD designs, as demonstrated in Figs.~\ref{fig:teaser} and~\ref{fig:edit}. These 3D shape editing results were obtained by MeshLab~\cite{meshlab}.

In addition, we can fit basic shapes (e.g., cuboids) to the abstracted convex shapes and export them to an OpenSCAD script, enabling programmatic editing of 3D shapes in CAD software~\cite{openscad}, as demonstrated in the top row in Fig.~\ref{fig:generation}. 

Last but not least, the abstracted 3D shapes obtained by DPA-Net can serve as easy-to-edit ``structural prompts'' ~\cite{sella2023spic,chen2023shaddr} for novel and detailed 3D shape generations. Here, we take the 3D generation method from SPiC-E~\cite{sella2023spic} as an example.} SPiC·E can transform primitive-based abstractions into highly expressive shapes according to text conditions. Taking one result from DPA-Net, for example, we produce three edited shapes from OpenSCAD as input guidance 3D shape, which are referred to as structural prompts in Fig.~\ref{fig:generation}. We input these guidance 3D shapes and different text conditions into SPiC-E and generate more detailed shapes. The results demonstrate that our method holds strong application potential for such promising downstream tasks.

\section{Conclusion}
\label{sec:conclu}

We present DPA-Net, a differentiable framework for learning structured 3D abstractions in the form of primitive assemblies from only few, e.g., three, RGB images captured at very different views. Our key innovation is the integration of differential primitive assembly into the NeRF architecture, enabling the prediction of occupancies to serve as opacity values for volume rendering. Without any 3D or shape decomposition supervision, our method can produce an interpretable, \rz{and subsequently editable}, union of convexes that approximates the target 3D object. \rz{Quantitative and qualitative evaluations on ShapeNet and DTU demonstrate superiority of DPA-Net over state-of-the-art alternatives.} The demonstrated applications further show that our editable 3D abstractions can serve as structural prompts and benefit other 3D generation tasks.

\rz{
Our current implementation makes use of GT camera poses. To alleviate performance degradations caused by estimated, noisy poses, existing methods for joint camera-scene optimization, e.g.,~\cite{truong2023sparf}, may be applied. As texture prediction is not a focus of our work, further fine-tuning (e.g., with a bias towards input views) and optimization are needed to improve rendering quality. Finally, assembly using only convexes is limiting. As shown in the supplementary, DPA-Net does not handle concave shapes well. Adding difference operations into the differentiable assembly is worth exploring.
}

%
%
\section*{Acknowledgement}
\fg{We thank all the anonymous reviewers and area chairs for their constructive comments. We also thank Kai Wang and Jianbo Ye both from Amazon for proofreading.}

\bibliographystyle{splncs04}
\bibliography{main}
\end{document}